\documentclass[11pt,a4paper]{article}
\usepackage[hyperref]{eacl2021}
\usepackage{times}
\usepackage{latexsym}

\usepackage{graphicx}
\usepackage{microtype}
\usepackage{xcolor}

\aclfinalcopy

\title{Few-shot Learning for Slot Tagging with Attentive Relational Network}

\author{Cennet Oguz$^{1}$ \hspace*{1.5cm} Ngoc Thang Vu$^{2}$\\
    $^1$Multilinguality and Language Technology, German Research Center for Artificial  Intelligence \\
    $^2$Institute for Natural Language Processing, University of Stuttgart \\
    \texttt{cennet.oguz@dfki.de} \hspace*{1.5cm}
    \texttt{thangvu@ims.uni-stuttgart.de}
}

\begin{document}
\maketitle
\begin{abstract}
Metric-based learning is a well-known family
of methods for few-shot learning, especially
in computer vision. Recently, they have been
used in many natural language processing applications but not for slot tagging. In this paper,
we explore metric-based learning methods in
the slot tagging task and propose a novel metric-based learning architecture - Attentive Relational
Network. Our proposed method extends relation networks, making them more suitable
for natural language processing applications
in general, by leveraging pretrained contextual
embeddings such as ELMO and BERT and by
using attention mechanism. 
The results on
SNIPS data show that our proposed method
outperforms other state of the art metric-based
learning methods.
\end{abstract}

\section{Introduction}
Neural networks have been successfully utilized in natural language processing (NLP) applications with a large amount of hand-labeled data whereas they suffer a persistent challenge of low-resource. The approach of learning with few samples, known as few-shot learning - a branch of meta-learning (\textit{learn to learn}) - has recently been popularized \cite{fei2006one,ravi2016optimization,vinyals2016matching,snell2017prototypical,sung2018learning} in computer vision. Recently, few-shot learning has also been applied to NLP tasks, e.g. natural language understanding \cite{Dou2019InvestigatingMA}, text classification \cite{jiang2018attentive,rios2018few,gao2019hybrid,Geng2019FewShotTC}, machine translation \cite{Gu2018MetaLearningFL} and relation classification \cite{obamuyide2019model}. 

In the slot tagging task, we aim at predicting task-specific values (e.g. artist, time) for slots (placeholders) in user utterances. 

\citet{oguz2020two} propose a two-stage modeling approach to exploit domain-agnostic features to tackle low-resource domain challenges. Besides, the other state of the art techniques e.g. based on external memory \cite{Peng2015RecurrentNN}, ranking loss \cite{vu2016bi}, encoder \cite{Kurata2016LeveragingSI}, and attention \cite{zhu2017encoder} have achieved promising results with a wide range of neural networks methods. 
 
However as many other NLP applications, the low-resource issue is a tremendous challenge for slot tagging in new domains, although labeled samples exist in related domains. Many studies have recently proposed to overcome this low-resource challenge using different techniques, e.g. multi-task modeling \cite{Jaech2016DomainAO}, adversarial training \cite{kim2017adversarial}, and pointer networks \cite{zhai2017neural}. In addition, studies like zero-shot learning has influenced the studies of the domain scaling problem for slots prediction \cite{Bapna2017TowardsZF}, eliminating the need of labeled examples for transferring reusable concepts \cite{Zhu2018ConceptTL,lee2019zero}, and conveying the domain-agnostic concepts between the intents \cite{Shah2019RobustZC} by exploiting label names and descriptions. Likewise, \cite{hou2020few} use label semantics within a few-shot classification method TapNet \citep{Yoon2019TapNetNN}.

We suggest using a small amount of annotated samples from different domains as training input instead of slot descriptions and slot names as in previos zero-shot \cite{Bapna2017TowardsZF,lee2019zero,Shah2019RobustZC} and few-shot \cite{hou2020few} slot tagging studies for two reasons: (1) The creation of slot descriptions needs qualified linguistic expertise and is thus expensive. (2) The relationship between slot names and the corresponding tokens is not constant. To give an example, the relationship between the '\textit{genre}' slot name and '\textit{drama}' token is hypernymic whereas the relationship between the '\textit{artist}' slot name and '\textit{Tarkan}' token is instance based. Hence, it may not be valid to learn only one function to represent the different relationships between names and tokens. 

In this paper, we provide a new experimental design where the slot tagging task needs to be solved for unseen slot labels. The experimental design mimics previous few-shot learning studies \cite{vinyals2016matching,snell2017prototypical,sung2018learning}. Thus, the existing data sources from different domains are used to learn meta-knowledge, whereas unseen labels from low-resource domains are used to evaluate the models. Furthermore, we propose a novel modeling approach - \emph{Attentive Relational Network}, inspired by \cite{sung2018learning,jiang2018attentive,Jetley2018LearnTP}, that leverages contextual embeddings such as ELMO and BERT and extends the previous relation networks \cite{sung2018learning} by learning to attend local and global features \cite{Jetley2018LearnTP}. Experimental results on SNIPS data show that the proposed model outperforms other few-shot learning networks. 

\section{Methods}

\subsection{Input}

\noindent \textbf{FastText} \cite{mikolov2018advances} is an approach to enrich the word vectors with a bag of character n-gram vectors.

\noindent \textbf{ELMo} \cite{Peters2018DeepCW} is a contextualized word representation methods. It concatenates the output of two LSTM independently trained on the bidirectional language modeling task and return the hidden states for the given input sequence.

\noindent \textbf{BERT} \cite{Devlin2019BERTPO} uses a bidirectional transformer model that is trained on a masked language modeling task. 
Because of WordPiece embeddings \cite{Wu2016GooglesNM}, there are different choices of presenting words. We use the first sub-token for representing the word as proposed in \cite{Devlin2019BERTPO}. Additionally, due to the structure of multiple successive layers, i.e., 24 layers and as suggested in \cite{oguz2020two}, we select \emph{10th}, \emph{11th}, \emph{12th}, and \emph{13th} as the focused layers on local context \cite{Clark2019WhatDB,Tenney2019WhatDY} for slot tagging. 
 
 \begin{figure}
    \centering
    \includegraphics[width=7.8cm,height=3cm]{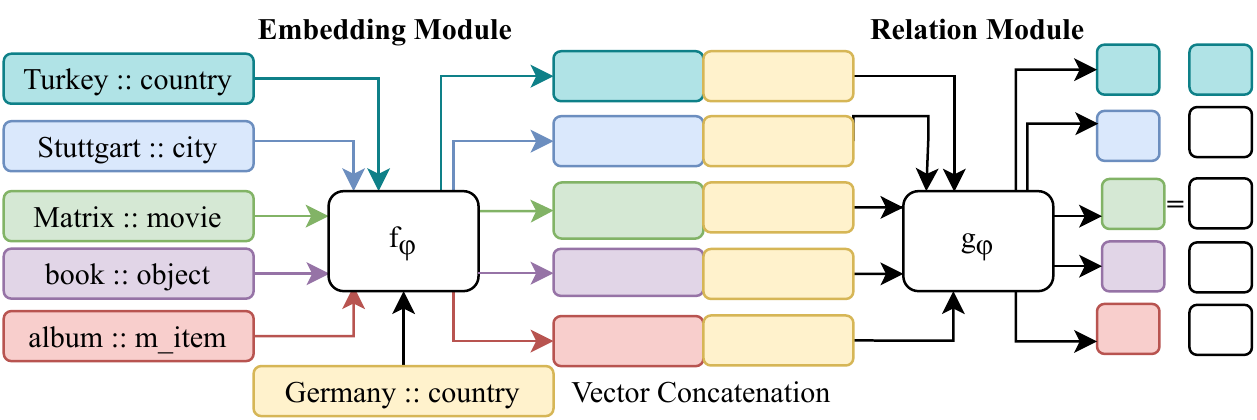}
    \caption{One-shot training example of slot tagging with Relation Network: Embedding Module extracts the feature vectors of each slot value whereas Relation Module calculates the relation scores between support samples and query. Then, the label of most relational value is assigned as a label of query.}
    \label{fig:one-shot}
\end{figure}
 
 \subsection{Meta-learning strategy}
 Despite the fact that the proposed methods differ in their learning strategies, episode-based training is the same in meta-training and meta-testing phases for proposed meta models as mentioned in \cite{Chen2019ACL}. For the purpose of applying episodic training in a robust way, we follow the proposed procedures in \newcite{vinyals2016matching,snell2017prototypical}. In the episodic training, each step - episode - is formed to compute gradients and update the model parameters. An episode consists of two components: support and query sets. To construct an episode, $C$ unique classes are randomly sampled,  and for each selected $C$ unique classes $K$ labeled examples randomly drawn for support $S=\{(x_i,y_i)\}^m_{i=1}$ and query $Q=\{(x_i,y_i)\}^m_{i=1}$ set, where $K > 1$ and $m=K*C$. The same episode composition strategy is applied in the meta-testing stage to evaluate the performance of the trained model over unseen classes.

\noindent \textbf{Meta-training.} The aim of this phase is to learn a meta learner that maps from a few labeled samples to a classifier. In each episode, meta-training employs a two-stage process: (1) the first stage implies producing the feature maps from the given input $S$ and $Q$, called embedding function $f_\phi(x)$ (2) the second stage is to make prediction conditioned on few labeled examples, $S$. More formally, we define an episode $E_{train}$ includes $S$ and $Q$ selected from train data,$D_{train}$ . Then, the model is trained to minimize the label prediction error in the $Q$  conditioned on $S$, i.e., $P_\theta(y_j|x_j,S)$, by utilizing the distance or relation metrics like  $distance/relation(f_\phi(x_j),f_\phi(x_i))$, as also shown with an example in Figure \ref{fig:one-shot}.

\noindent \textbf{Meta-testing.} In this phase, we test the performance of trained meta-learner on unseen labels by following the same steps in meta-training phase. An episode $E_{test}$ with $S$ and $Q$ is formed by randomly selecting from test data, $D_{test}$. The over all accuracy is computed by averaging the test episodes, 
$ acc = \frac{1}{||E_{test}||} \sum_i E_{test}$.

We define $C = 5$ in meta-training and meta-testing stages except the meta-testing stage of \emph{SearchCreativeW.} domain with $C = 3$ because \emph{SearchCreativeW.} domain has only three slots. 
We train all the model within 10,000 episodes, and evaluate with 1000 test episode after every 500 steps of total train episodes.

\subsection{Models}
 \noindent We focus on three \emph{metric-based learning} few-shot learning methods such as Matching Networks \cite{vinyals2016matching}, Prototypical Network \cite{snell2017prototypical}, and Relation Network \cite{sung2018learning}. Each network consists of two consecutive modules. The first module, called embedding function,  focuses on the learning of the transferable embeddings for support and query samples. The second module is the classifier which identify the corresponding classes over the defined metric scores, e.g., distance and relation.
 
 \noindent \textbf{Matching Networks} (MatchingNets) compare the cosine distance between the
query feature and each support feature, and computes average cosine distance for each class. 

\noindent \textbf{Prototypical Networks} (PrototypicalNets) compare the Euclidean distance between query features and the class mean of support features. 

\noindent \textbf{Relation Networks} (RelationNets) propose a learnable non-linear relation module to output the relation scores over element-wise sum of each support and query features.

\noindent In one-shot scenario, MatchingNets and PrototypicalNets could be interpreted as identical, RelationNets differs with the relation module in order to calculate the relation score.

\begin{figure}
    \centering
    \includegraphics[width=7cm,height=4.5cm]{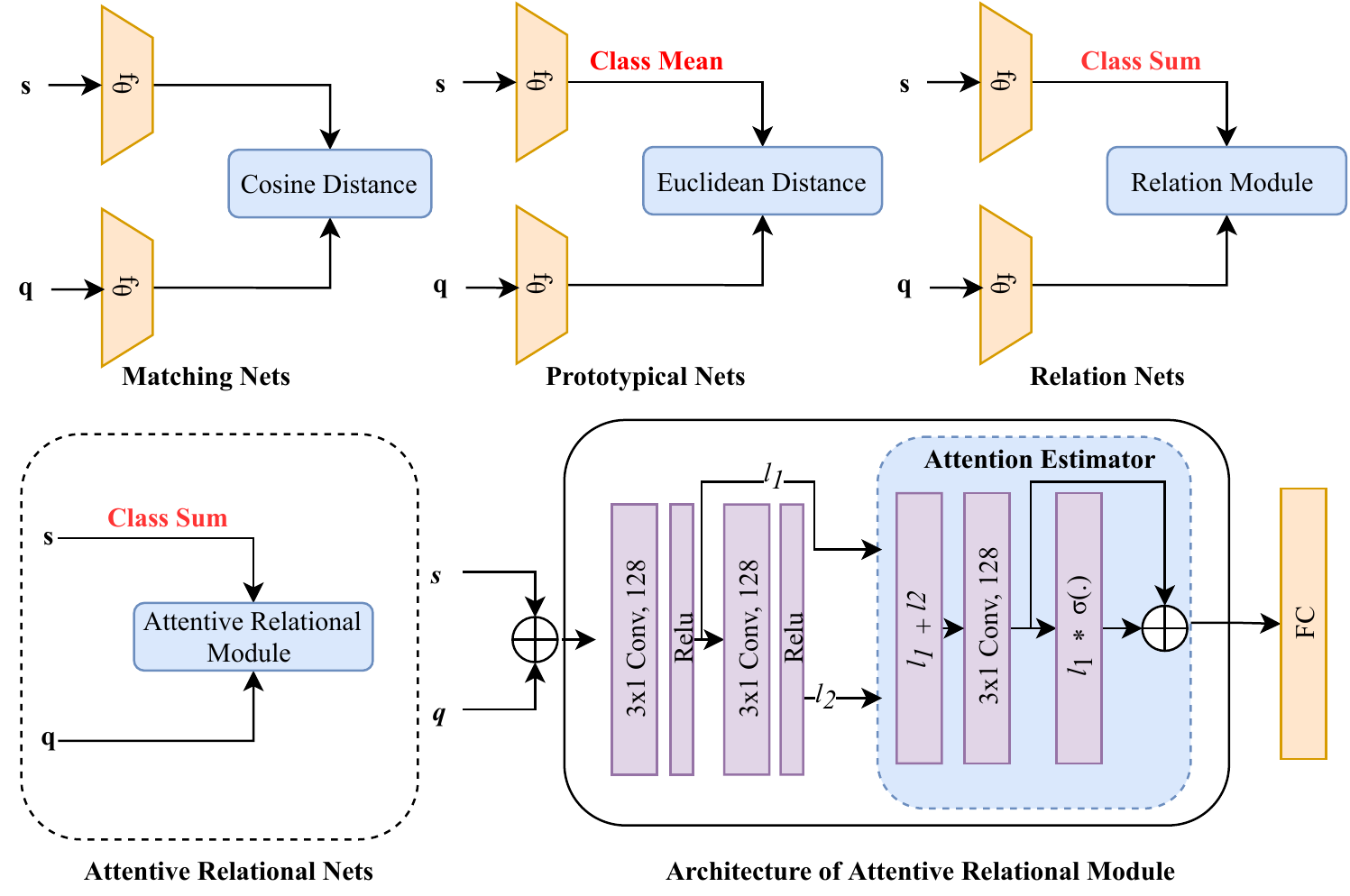}
    \caption{Few-shot learning models: Matching Nets compare each support sample with query in order to calculate the distance metric, Prototypical Nets rely on class mean of support set, and Relation Nets are based on class sum of support samples. $f_\theta$ represent the embedding module of each network and Attentive Relational Nets eject embedding module. Support sets are represented with $s$ whereas $q$ denotes the query.}
    \label{fig:metamodels}
\end{figure}

\subsubsection{Attentive Relational Networks} 
\noindent We propose a novel metric-learning approach - Attentive Relational Networks (AttentiveRelationalNets) that
highlight the relevant, and suppress the misleading between support and query samples.  AttentiveRelationalNets address the few-shot classification problem by utilizing \emph{learn to compare based on attention} insight. This  can be seen as extending the strategy
of \newcite{sung2018learning} to include a learnable attention module. A trainable attention module, inspired from \newcite{Jetley2018LearnTP}, is added to incorporate the relation module of RelationNets. Besides, we make use of pretrained (contextual) embeddings since they have the proven strength on feature extraction for linguistics items instead of using embedding module, Figure \ref{fig:metamodels}.  

\noindent For AttentiveRelationalNets, as shown in Figure \ref{fig:metamodels}, we implement two convolution blocks as it is in RelationNets with residual connection, as proposed in \newcite{he2016deep}.
Then, the convolution blocks produce local descriptors, i.e., $l_1$ and $l_2$, as the output of activation function and pass them to the attention estimator in order to find the global $g$ feature vector.

\noindent In order to compute the compatibility function, we define a convolution function with the input of two local features to an addition operation, $c= \langle u, l_1 + l_2 \rangle $. Here, $u$ represents the universal set of features relevant to the $s$ and $q$ pairs in the object categories. We normalize the compatibility scores by using sigmoid operation, $a = \sigma{(c)} $. Then, the global feature vector is assessed by element-wise weighted average, i.e, $g = l_1 * a$. Afterwards, we concatenate the global features $g$ with learned compatibility scores $c$ as the input of the linear classifier which eventually produces a scalar in range of 0 to 1 representing the similarity between $s$ and $q$ , which is called relation score $r$. We define \emph{mean square error}, as proposed in RelationNets, as the objective function of our model.
 
 \subsection{Evaluation}
As we use the same implementation details for meta-training and meta-testing stages, we also evaluate the performance by few-shot classification accuracy following previous studies few-shot learning \cite{vinyals2016matching,snell2017prototypical,sung2018learning} with a small change: since the meta-learning approaches are fast learning methods, we present the average accuracies of training epochs instead of presenting the best accuracy.

\begin{figure}
    \centering
    \includegraphics[width=7cm,height=4.5cm]{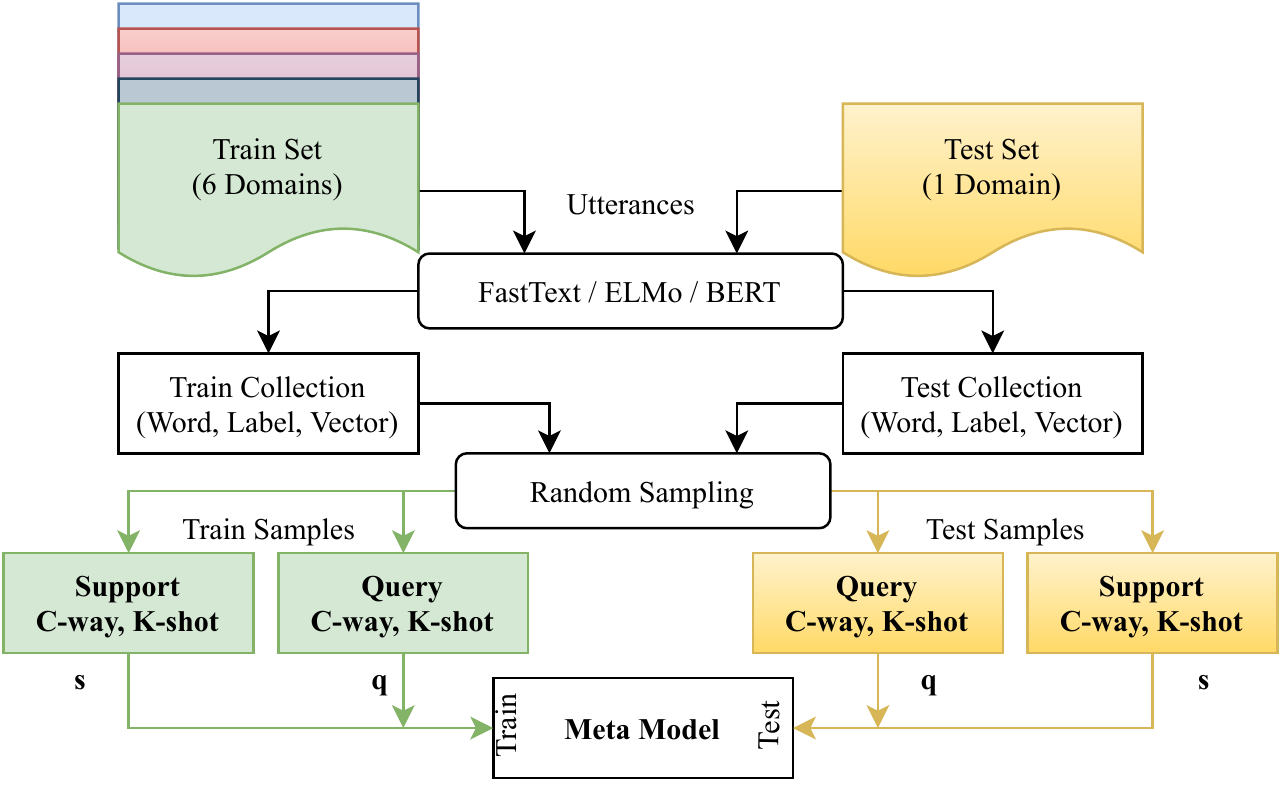}
    \caption{The schema of few-shot data construction for train and test episodes. 6 different domains are used for training phase whereas one different domain is  used for test. Green path indicates the training support and query samples while yellow path represents the samples of test episodes. Train and test collections include $N$ number of values of each slot. Random sampling function draws $K$ samples for each of $C$ slots.}
    \label{fig:dataflow}
\end{figure}

\section{Settings}

\subsection{Resources}
In our study we address the few-shot learning approaches to recognize novel slot categories with very few examples from  a new domain. In order to provide a deep experimental analysis of proposed networks and language models and to compare our model among each other, we set various experimental scenarios with different data and different $K$-shot sizes. Hence, we utilize the SNIPS dataset \cite{Coucke2018SnipsVP} as a base dataset in our experiment. SNIPS is a SLU dataset of crowd-sourced user utterances with 39 slots and 7 intents. Thus, it is a well-categorized dataset which include tasks in domains, which makes the setup more realistic; learn to learn on a bunch of domains and test on new domains. We split SNIPS with the purpose of creating a single-domain dataset. We combine the originally divided training, testing, and development sets and separate them into domain sets in order to create new train and test data. 
 
 \subsection{Few-shot Data Construction}
  Meta-learning models aim to learn from the training tasks, i.e. the train label space is disjoint with test label space and the trained model evaluated on unseen classes. Therefore, we utilize other domain data as the training set whereas the models are evaluated by using the current domain. Thus, we created 7 different sets contain a train which consists of 6 different domains as well as a test set which includes only one test domain.
  
As can be seen from Figure \ref{fig:dataflow}, we aggregate the six different domain data for the training set, whereas the remain one domain is used for testing, aiming at evaluating the performance of models on unseen classes per domain. Then, we convert the train and test sets to \emph{train} and \emph{test} collection that contain triplets in order to mimic the same data organization in the previous meta-learning studies. The triplet consists of three items: token, label, vector. The vector of the corresponding token is produced by using different (contextual) embeddings from randomly selected sentences for each label from the train and test set separately. Thus, the train collection is formed with the triplets from the different domain slots, whereas the test collection includes only the triplets of labels from the corresponding domain.

To investigate the efficiency of different models according to data availability in the few-shot setting, we experiment with three data collections sizes of slot values: 50, 100, and 200. Note that the collection size controls the total number of values that can be seen for each slot during training. In the meta-testing stage, we only use the test collection with size of 200 to be able to keep the comparative analyses under control. Furthermore, we examine different numbers of $K$ shot, namely 5-shot, 10-shot, and 15-shot.

 \begin{table*}[!h]
\centering \caption{ Few-shot slot tagging on SNIPS data. Results are accuracies averaged of three different slot value sizes (50, 100, and 200) with different $K$s (5, 10, and 15).}
\resizebox{455pt}{!}{\begin{tabular}{|l|l|
                                   c|c|c|c|c|c|c|c|c|
                                   c|c|c|c|c|c|c|c|c|
                                   c|c|c|c|c|c|c|c|c|}
\hline
  & \multicolumn{9}{|c|}{\small{FastText}}
  & \multicolumn{9}{|c|}{\small{ELMo}}
  & \multicolumn{9}{|c|}{\small{BERT}}
\\

\hline
  & \multicolumn{3}{|c|}{\small{MatchingNet}}
  & \multicolumn{3}{|c|}{\small{PrototypicalNet}} 
  & \multicolumn{3}{|c|}{\small{RelationNet}}
  
  & \multicolumn{3}{|c|}{\small{MatchingNet}}
  & \multicolumn{3}{|c|}{\small{PrototypicalNet}} 
  & \multicolumn{3}{|c|}{\small{RelationNet}}
  
  & \multicolumn{3}{|c|}{\small{MatchingNet}}
  & \multicolumn{3}{|c|}{\small{PrototypicalNet}} 
  & \multicolumn{3}{|c|}{\small{RelationNet}}
  
  \\
\hline
 \small{Domain / K-shot} 
 & \small{5} & \small{10} & \small{15} 
 & \small{5} & \small{10} & \small{15} 
 & \small{5} & \small{10} & \small{15}
                
 & \small{5} & \small{10} & \small{15} 
 & \small{5} & \small{10} & \small{15} 
 & \small{5} & \small{10} & \small{15}
                
 & \small{5} & \small{10} & \small{15} 
 & \small{5} & \small{10} & \small{15} 
 & \small{5} & \small{10} & \small{15}\\
\hline
        \small{AddToPlaylist} 
        & \small{20.4} & \small{20.5} & \small{20.5}
        & \small{67.4} & \small{67.6} & \small{67.6}
        & \small{65.9} & \small{70.4} & \small{71.7}
        & \small{65.4} & \small{71.7} & \small{73.9}
        & \small{72.1} & \small{71.8} & \small{71.1}
        & \small{70.0} & \small{75.1} & \small{\textbf{76.3}}
        & \small{65.9} & \small{70.6} & \small{73.2}
        & \small{70.8} & \small{71.5} & \small{71.4}
        & \small{69.9} & \small{74.0} & \small{74.3} \\ 
        
       \small{PlayMusic}
        & \small{20.9} & \small{20.8} & \small{20.7}
        & \small{68.9} & \small{68.6} & \small{69.1}
        & \small{68.8} & \small{72.2} & \small{73.7}
        & \small{67.3} & \small{73.7} & \small{75.9}
        & \small{73.5} & \small{72.2} & \small{71.5}
        & \small{72.2} & \small{76.5} & \small{\textbf{78.0}}
        & \small{64.8} & \small{70.9} & \small{73.3}
        & \small{67.4} & \small{68.2} & \small{67.7}
        & \small{67.7} & \small{71.6} & \small{73.0} \\ 
                                       
        \small{BookRestaurant}
        & \small{23.5} & \small{23.8} & \small{24.0}
        & \small{75.0} & \small{74.5} & \small{74.3}
        & \small{78.7} & \small{82.5} & \small{83.6}
        & \small{71.3} & \small{76.8} & \small{78.9}
        & \small{81.2} & \small{79.5} & \small{77.5}
        & \small{82.9} & \small{85.9} & \small{\textbf{86.5}}
        & \small{68.4} & \small{72.7} & \small{74.8}
        & \small{75.8} & \small{76.5} & \small{76.8}
        & \small{78.4} & \small{81.8} & \small{83.7} \\ 
                      
        \small{GetWeather}
        & \small{20.7} & \small{25.7} & \small{20.7}
        & \small{76.6} & \small{75.9} & \small{75.7}
        & \small{79.5} & \small{84.2} & \small{85.3}
        & \small{75.6} & \small{82.2} & \small{84.8}
        & \small{80.1} & \small{78.5} & \small{77.9}
        & \small{79.8} & \small{85.1} & \small{\textbf{87.0}}
        & \small{75.3} & \small{81.7} & \small{84.3}
        & \small{78.5} & \small{78.9} & \small{79.1}
        & \small{83.6} & \small{86.2} & \small{85.8} \\ 
                      
        \small{RateBook}   
        & \small{34.5} & \small{35.1} & \small{35.2}
        & \small{87.4} & \small{88.3} & \small{87.2}
        & \small{89.8} & \small{92.2} & \small{93.9}
        & \small{83.8} & \small{89.3} & \small{90.6}
        & \small{90.2} & \small{90.2} & \small{89.6}
        & \small{90.0} & \small{93.9} & \small{\textbf{95.1}}
        & \small{88.0} & \small{92.0} & \small{93.7}
        & \small{92.8} & \small{93.6} & \small{93.2}
        & \small{92.0} & \small{94.1} & \small{93.9} \\ 
                      
        \small{SearchCreativeW.} 
        & \small{37.8} & \small{37.9} & \small{38.2}
        & \small{78.1} & \small{78.2} & \small{78.2}
        & \small{82.8} & \small{85.9} & \small{86.3}
        & \small{82.8} & \small{86.4} & \small{88.6}
        & \small{85.8} & \small{85.7} & \small{83.8}
        & \small{89.7} & \small{92.1} & \small{\textbf{92.9}}
        & \small{79.7} & \small{84.4} & \small{87.1}
        & \small{81.4} & \small{82.0} & \small{81.8}
        & \small{86.0} & \small{90.3} & \small{91.3} \\ 
                      
        \small{FindScreeningE.}  
        & \small{21.2} & \small{21.3} & \small{21.4}
        & \small{73.4} & \small{73.5} & \small{73.7}
        & \small{78.8} & \small{83.1} & \small{84.6}
        & \small{80.7} & \small{87.7} & \small{90.9}
        & \small{83.5} & \small{81.6} & \small{81.0}
        & \small{86.8} & \small{89.9} & \small{\textbf{91.7}}
        & \small{78.6} & \small{86.4} & \small{89.5}
        & \small{78.0} & \small{78.1} & \small{78.3}
        & \small{81.3} & \small{86.8} & \small{87.9} \\ 
\hline
\end{tabular}} 
  \label{previous_results}
\end{table*}
 {\small
\begin{table}[!t]
\caption{Few-shot slot tagging on SNIPS data with Attentive Relational Networks. Results are accuracies averaged of three different slot value sizes (50, 100, and 200) with different $K$s (5, 10, and 15).}
\centering
\resizebox{225pt}{!}{\begin{tabular}{ |l|l|c|c|c|c|c|c|c|c|c|}
\hline
  & \multicolumn{3}{|c|}{\small{FastText}} & \multicolumn{3}{|c|}{\small{ELMo}} &
  \multicolumn{3}{|c|}{\small{BERT}}\\
\hline
 \small{Domain} &  \small{5} & \small{10} & \small{15} & \small{5} & \small{10} & \small{15} & \small{5} & \small{10} & \small{15} \\
\hline
        \small{AddToPlaylist} 
        & \small{63.6} & \small{66.1} & \small{68.3}
        & \small{71.9} & \small{75.8} & \small{77.7} 
         & \small{72.6} & \small{76.6} & \small{\textbf{78.5}}\\ 
        
       \small{PlayMusic}  
       & \small{68.5} & \small{71.8} &\small{73.2}
       & \small{74.7} & \small{77.7} & \small{79.2} 
        & \small{72.8} & \small{77.0} & \small{\textbf{78.7}}\\
                                       
        \small{BookRestaurant}   
        & \small{78.9}  & \small{82.0}  & \small{83.1}
        & \small{84.2} & \small{87.2} & \small{\textbf{87.6}} 
         & \small{82.7} & \small{86.0} & \small{87.5}\\
                      
        \small{GetWeather} 
        & \small{79.0} &\small{82.7} &\small{84.5}
       & \small{83.3} & \small{87.1} & \small{88.5} 
        & \small{84.9} & \small{89.5} & \small{\textbf{90.2}}\\  
                      
        \small{RateBook }   
        & \small{88.1} &\small{90.9} &\small{91.4}
        & \small{92.3} & \small{94.8} & \small{95.6} 
         & \small{94.8} & \small{96.5} & \small{\textbf{96.7}}\\
                      
        \small{SearchCreativeW.} 
        & \small{81.7} &\small{84.4} & \small{85.1}
        & \small{90.8} & \small{93.0} & \small{\textbf{93.2}}
         & \small{88.2} & \small{91.5} & \small{92.6} \\
                      
        \small{FindScreeningE.}  
        & \small{76.9} & \small{81.2} & \small{82.1}
        & \small{87.3} & \small{90.8} & \small{\textbf{92.3}}
        & \small{82.9} & \small{87.0} & \small{88.3}\\  
\hline
\end{tabular}} 
  \label{attentive_results}
\end{table}}
\section{Results and Analysis}
\subsection{Different Network Architectures}
Table \ref{previous_results} shows the performance in comparison with state of the art  metric-based learning models on slot tagging task with different (contextual) embeddings. As can be seen from the scores, RelationNets with ELMo embeddings constantly give the best results,  whereas MatchingNets present the lowest scores for each embedding variance. We assume that learning from the distance or relation scores with individual support and query samples instead of the class sum, i.e., as it is in PrototypicalNets, or class mean, i.e., as it is in RelationNets, decrease the learning performance.

Furthermore, Table \ref{attentive_results} presents the results from AttentiveRelationalNets with different embeddings methods and demonstrates that our proposed model with ELMo and BERT outperforms the previous models consistently. Additionally, AttentiveRelationalNets significantly improve the results with BERT from the previous experiments in Table \ref{previous_results}. When the success of AttentiveRelationalNets is examined extensively,
it seems that our proposed model gives better results on slot labels that categorize common nouns, while provides relative competitive results for proper nouns. Note that proper nouns, in general, seem to be challenging for all setups. 
In addition, RelationalNets outperform AttentiveRelationalNets with FastText along with the explanation that embedding function is still effective with average embeddings. However, as opposed to RelationNets, Attentive Relational Networks do better classification overall because of trainable attention. Since trainable attention highlights the relevant features between the slot values labeled with the same slot, whereas it suppresses the misleading them. In an other word, slot local features are able to be more informative for the model while the global features are suppressed.

\subsection{Different Contextual Embeedings}
Table \ref{attentive_results} shows that ELMo and BERT have comparable performance, with BERT slightly better on most tasks: ELMo, however, scores higher on  \emph{FindScreeningE.} consistently with AttentiveRelationalNets and all different $K$s. 
Although embedding modules are presented as a feature extraction method for inputs according to distance or relational score, the significant performance gap between FastText and contextualized embeddings shows that the contextualized features outperform the embedding module of few-shot classification models. On the other hand, when we compare FastText embeddings with contextualized word vectors in Table \ref{attentive_results}, the lower results can be seen. Additionally, when FastText features are compared between the results of RelationNets and AttentiveRelationalNets, we observe that RelationNets outperform AttentiveRelationalNets.

\noindent We further look at the wrong predictions in order to understand the reason for the success of ELMo on \emph{FindScreeningE.} and observe that ELMo shows high performance on the labels of proper nouns such as city and location and the labels like \emph{object\_type} and \emph{movie\_type}. However, BERT demonstrates high performance on proper nouns such as artist, album, hence it outperforms ELMo on \emph{AddToPlaylist} and \emph{PlayMusic} domains. In addition, the improvement of BERT with AttentiveRelationalNets mostly relies on the increase of the accuracy of overall labels, but especially the slot labels that contain common nouns.

\subsection{Different Collection and Shot Sizes} AttentiveRelationalNets demonstrate a linear correlation between the increase of performance and the increase of collection size. In addition, the increase of shot size mostly shows improvement in overall results, apart from PrototypicalNets which show their highest result with 10-shot.

\section{Conclusion}
We presented a deep analysis with a wide variety of few-shot learning methods and pretrained (contextual) embeddings for slot tagging. Furthermore, we proposed a novel architecture that leverages attention mechanism attending both, local and global features of given support samples. Experimental results on SNIPS dataset show that a) pretrained contextual embeddings contributed to high performance and b) our proposed approach consistently outperformed other methods in all setups.

\bibliography{anthology,eacl2021}
\bibliographystyle{acl_natbib}

\end{document}